\documentclass{sjtu_sai}

\usepackage{amsmath,amsfonts,bm}

\def\eqref#1{equation~\ref{#1}}

\def\1{\bm{1}}

\DeclareMathAlphabet{\mathsfit}{\encodingdefault}{\sfdefault}{m}{sl}
\SetMathAlphabet{\mathsfit}{bold}{\encodingdefault}{\sfdefault}{bx}{n}

\usepackage{float}
\usepackage{hyperref}
\usepackage{url}
\usepackage{graphicx}
\usepackage{wrapfig}
\usepackage{booktabs}
\usepackage{multirow}
\usepackage{amsmath}
\usepackage{xcolor}
\titleformat{\paragraph}[runin]{\normalfont\normalsize\bfseries}{}{0pt}{}
\titlespacing*{\paragraph}{0pt}{1.5ex plus .3ex}{0.7em}
\DeclareMathAlphabet{\mathcal}{OMS}{cmsy}{m}{n}

\title{Acting through Registers}
\renewcommand{\authorformat}[2][]{\mbox{\authorfont\bfseries\rmfamily #2$^{#1}$}}
\author[1,2,3]{Yiwei Zhang}
\author[1,3]{Yantai Yang}
\author[2,\dagger]{Jin Gao}
\author[3]{Jin Lou}
\author[3]{Andong Chen}
\author[3]{Yufei Liu}
\author[3]{Boyang Xing}
\author[3]{Yuchen Zhu}
\author[2]{Weiming Hu}
\author[1,\dagger]{Zhipeng Zhang}
\affiliation[1]{AutoLab, School of Artificial Intelligence, Shanghai Jiao Tong University}
\affiliation[2]{Institute of Automation, Chinese Academy of Sciences}
\affiliation[3]{Humanoid Robot (Shanghai) Co., Ltd.}
\renewcommand{\affiliationlist}{%
\affiliationformat[1]{AutoLab, School of Artificial Intelligence, Shanghai Jiao Tong University}\par
\affiliationformat[2]{Institute of Automation, Chinese Academy of Sciences}\par
\affiliationformat[3]{Humanoid Robot (Shanghai) Co., Ltd.}\par}

\contribution[\dagger]{Corresponding authors}

\abstract{

World Action Models (WAMs) use generative visual representations to condition robot actions, 
yet how these representations organize action-relevant information remains poorly understood. 
We identify persistent high-norm spatial outliers in WAM representations and find that their 
action relevance persists even after normalization and K/V projection attenuate their extreme magnitude. 
Motivated by this observation, we introduce RegisterWAM, which adds learnable non-spatial registers 
to the world model and uses them directly for action conditioning. Registers alone do not fully 
eliminate spatial outliers; pairing them with observation flow-matching (FM) loss supervision 
suppresses spatial outliers while concentrating high-norm responses in registers. 
Beyond this reorganization, observation FM loss supervision also improves action learning without registers, 
motivating an additional noised-observation FM objective that learns current-scene recovery from corrupted inputs. 
The resulting model achieves 81.64\% success on LIBERO-Plus without additional embodied pretraining 
data or foundation models, reaches 93.0\% average success on RoboTwin 2.0, 
and improves real-world manipulation performance.
}
\renewcommand{\checkdataformat}[2][]{\makebox[\linewidth][c]{\small{\checkdatafont\sffamily\bfseries #1:} #2}}
\checkdata[Code]{\url{https://github.com/Z1zyw/RegisterWAM}}
\newcommand{\institutionlogos}{%
  \includegraphics[height=9mm]{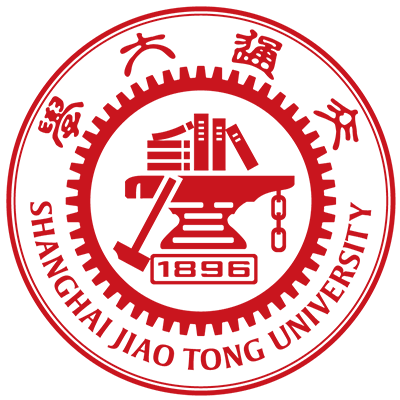}\hspace{5mm}%
  \includegraphics[height=9mm]{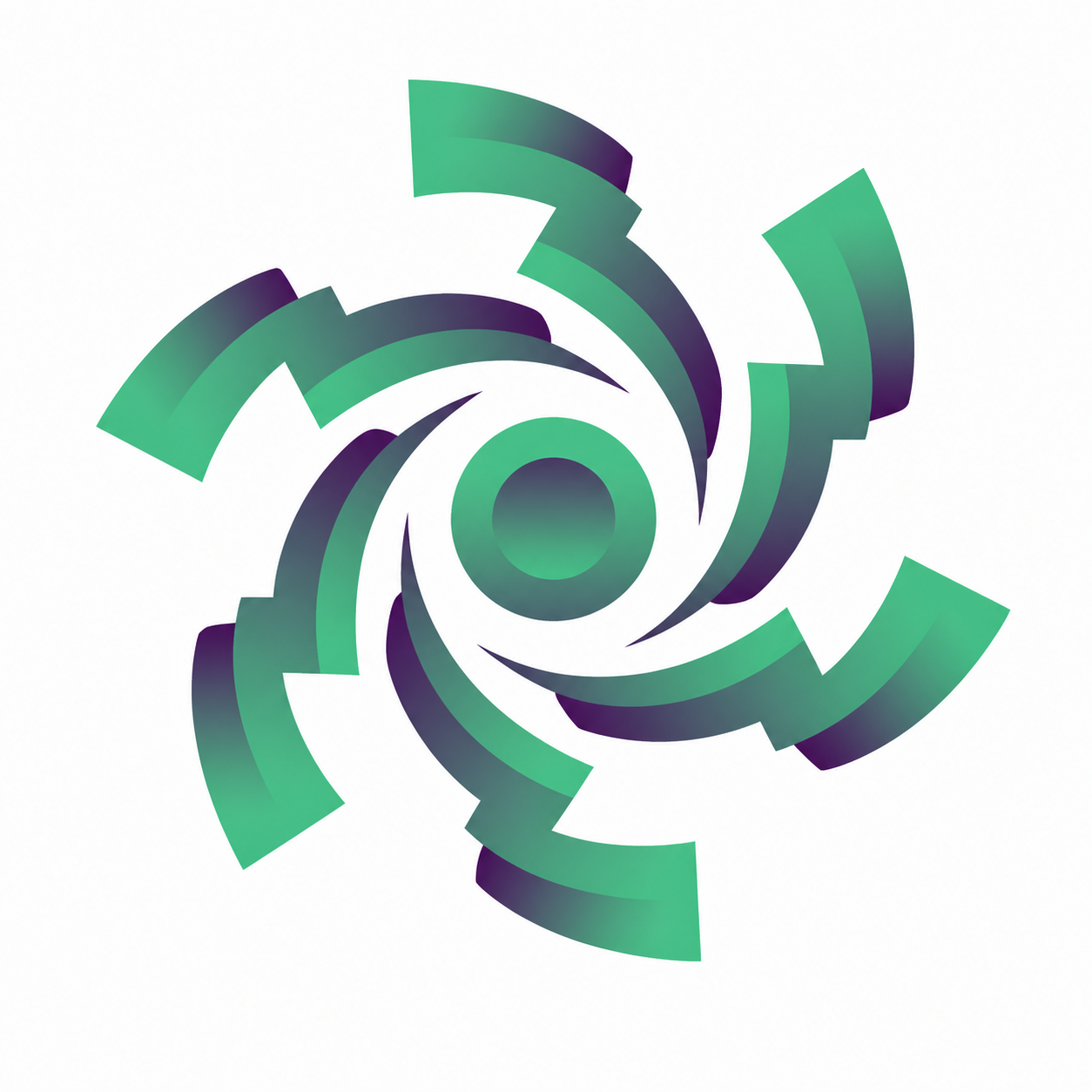}\hspace{5mm}%
  \includegraphics[height=9mm]{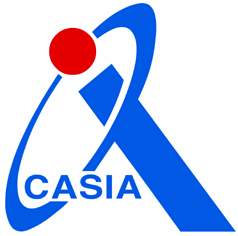}\hspace{5mm}%
  \raisebox{-1.2mm}{\includegraphics[height=11mm,trim=100.8bp 0 136.8bp 0,clip]{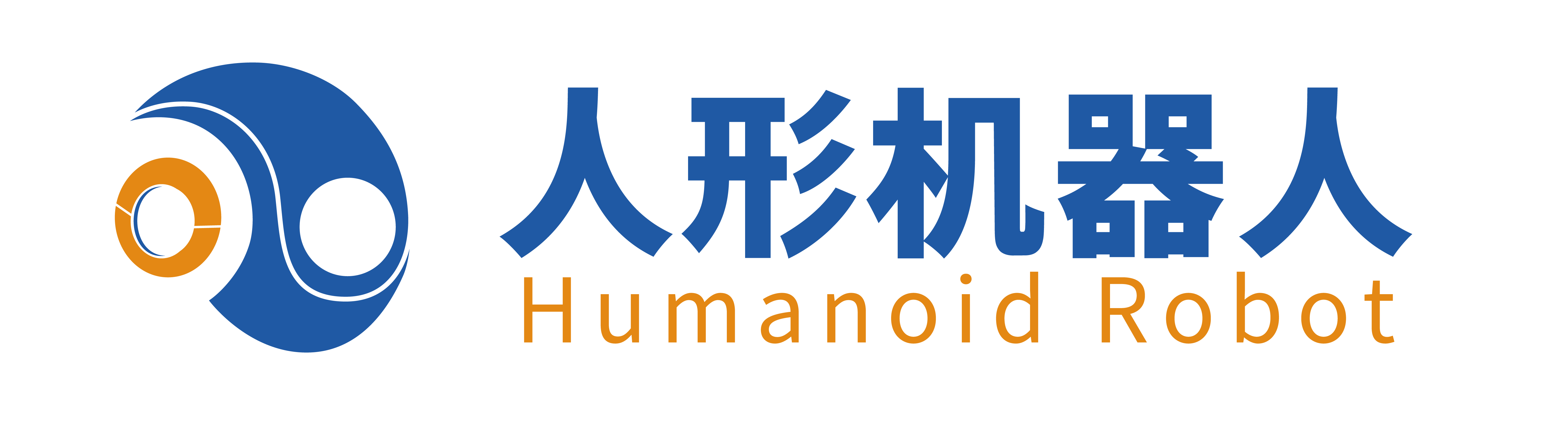}}%
}
\fancypagestyle{firststyle}{\fancyhf{}\fancyfoot[C]{\thepage}}
\makeatletter
\renewcommand{\@toptitlebar}{%
  \noindent\makebox[\linewidth][l]{\institutionlogos}\par
  \vskip 2mm
  {\color{saiblue}\hrule height 0.5pt}%
  \vskip 6mm
}
\makeatother
\hypersetup{pdftitle={Acting through Registers},pdfauthor={Yiwei Zhang, Yantai Yang, Jin Gao, Jin Lou, Andong Chen, Yufei Liu, Boyang Xing, Yuchen Zhu, Weiming Hu, Zhipeng Zhang}}
\begin{document}
\maketitle

\section{Introduction}
\label{sec:intro}

World Action Models (WAMs) adapt generative video models
to jointly model visual dynamics and robot actions
\citep{cheang2024gr2generativevideolanguageactionmodel,liang2025videogenerators,zhu2025unifiedworldmodelscoupling,kim2026cosmos}.
The video backbone encodes the current scene and its possible evolution,
and the resulting representations are used to condition an action model
\citep{hu2024video,ge2025,pai2025mimicvideo}.
This formulation allows robot policies to benefit from large-scale video pretraining
and predictive representations of physical dynamics~\citep{yuan2026fastwam,zhang2026imagewam,ma2026fasterwam,zhao2026fasterwam}.
However, how these video representations behave inside WAMs,
and how such behavior relates to action generation,
remain poorly understood.

\begin{figure*}[t]
\centering
\begin{minipage}{0.90\textwidth}
\centering
\begin{minipage}[t]{0.328\linewidth}
\centering\small\textbf{Pretrained video models}\\[1.5pt]
\includegraphics[width=\linewidth]{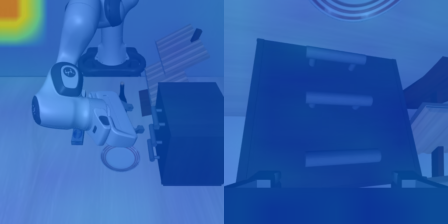}\\[-1pt]
{\scriptsize (a) Wan2.2: hidden-state norm}
\end{minipage}\hfill
\begin{minipage}[t]{0.328\linewidth}
\centering\small\textbf{FastWAM: Sample 1}\\[1.5pt]
\includegraphics[width=\linewidth]{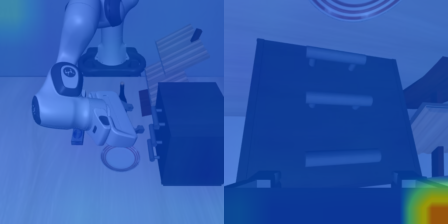}\\[-1pt]
{\scriptsize (c) Hidden-state norm}
\end{minipage}\hfill
\begin{minipage}[t]{0.328\linewidth}
\centering\small\textbf{FastWAM: Sample 2}\\[1.5pt]
\includegraphics[width=\linewidth]{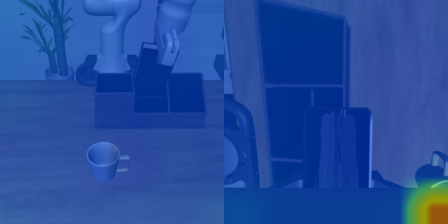}\\[-1pt]
{\scriptsize (e) Hidden-state norm}
\end{minipage}

\vspace{2pt}
\begin{minipage}[t]{0.328\linewidth}
\centering
\includegraphics[width=\linewidth,height=0.5\linewidth]{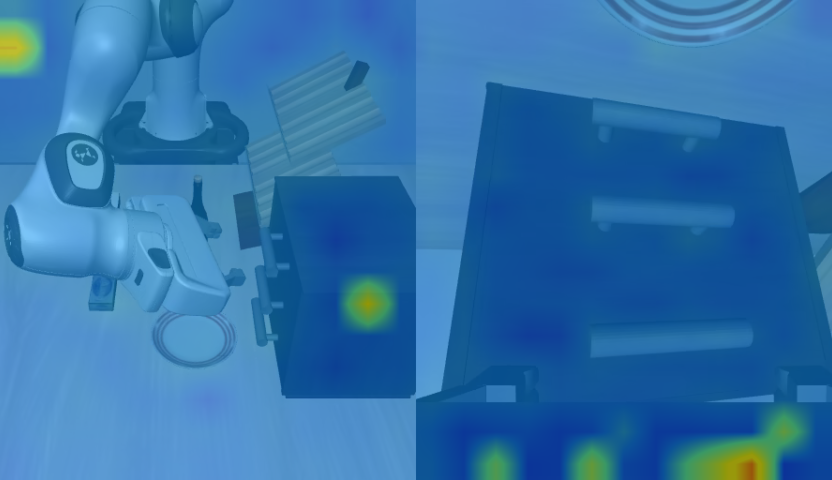}\\[-1pt]
{\scriptsize (b) Cosmos3-Nano: hidden-state norm}
\end{minipage}\hfill
\begin{minipage}[t]{0.328\linewidth}
\centering
\includegraphics[width=\linewidth]{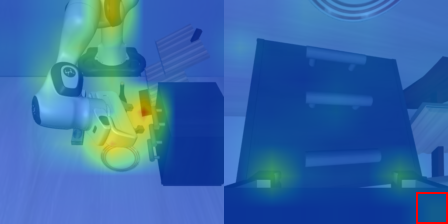}\\[-1pt]
{\scriptsize (d) Masked-token action sensitivity}
\end{minipage}\hfill
\begin{minipage}[t]{0.328\linewidth}
\centering
\includegraphics[width=\linewidth]{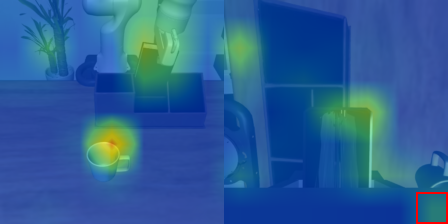}\\[-1pt]
{\scriptsize (f) Masked-token action sensitivity}
\end{minipage}
\end{minipage}
\caption{\small\textbf{Spatial outliers and action relevance.} Hidden-state norms and masked-token action sensitivity across video models and FastWAM.}
\label{fig:hidden_action_intro}
\vspace{-0.6em}
\end{figure*}

To investigate how video representations may support action generation, we examine how spatial tokens evolve across the layers of the video backbone. A striking pattern emerges. In the middle-to-late layers, a small number of tokens develop exceptionally large hidden-state norms, approximately an order of magnitude greater than those of the remaining tokens (Fig.~\ref{fig:hidden_action_intro}). High-norm outliers are not entirely absent from pretrained video models, but they are typically weaker and less consistent. In the WAM studied here, they become substantially more pronounced and persist across different observations. Their amplification in a model adapted for action generation points to a potentially meaningful role, yet their function remains unclear. Then, \textit{are these high-norm outliers merely numerical artifacts, or do they participate in the generation of robot actions?}

To answer this question,
we trace the outliers through the visual K/V cache
that connects the Video DiT to the Action DiT. Interestingly, we observed that, despite their extreme hidden-state norms, normalization and projection substantially attenuate the magnitude gap in the cached keys and values. The unusually large norms are therefore not directly propagated to the action model. Their functional relevance, however, persists. Token-level interventions on the K/V cache, implemented by masking visual positions from action attention, measure the resulting change in predicted actions, which we term action sensitivity. The entries associated with these outliers retain non-trivial action sensitivity (Fig.~\ref{fig:hidden_action_intro}). Their action sensitivity is not necessarily dominant, as tokens aligned with task-relevant objects often produce larger action changes. What distinguishes the outlier tokens is instead the consistency of their action relevance across observations, even when their spatial locations contain little task-relevant information. Taken together, these results suggest that high-norm outliers are not simply numerical artifacts. Rather, they provide a visible signature of spatial tokens whose representational roles have become partially decoupled from their nominal spatial locations.

The partial decoupling between token function and spatial location suggests a possible architectural tension. Although visual tokens are organized around local image regions, the world model may also need to represent information that is useful for action generation but not naturally associated with any single location. In the absence of dedicated non-spatial capacity, some spatial tokens may be recruited to serve such roles, with their high norms providing an observable signature of this functional shift. This hypothesis motivates a natural architectural question: \textit{can these representation roles be better accommodated by tokens that have no spatial identity?}

To investigate this possibility, we propose \emph{RegisterWAM}, which introduces a small set of dedicated non-spatial tokens, termed \emph{registers}, into the world model. Unlike visual tokens, registers have no explicit correspondence to image regions. They interact with spatial tokens throughout the world model and directly condition the action model, providing representation capacity that is not tied to any particular spatial location. Registers alone do not fully eliminate spatial outliers. Although they suppress intermediate-layer peaks, outliers remain in the final layers (Appendix~\ref{app:additional_outliers}). We therefore introduce observation flow-matching (FM) loss supervision on the dense clean-observation tokens, encouraging them to remain predictive of visual observations while leaving registers free to capture information that need not be spatially localized. Under this joint design, high-norm outliers disappear from the spatial tokens and become concentrated in the registers. Thus, the observable outlier behavior can be separated from spatial representations when dedicated non-spatial capacity is paired with explicit observation FM loss supervision. This reorganization is accompanied by improved policy performance. Register-only and raw-plus-register conditioning achieve broadly comparable performance across the settings in Figure~\ref{fig:register_count_analysis}. We adopt register-only conditioning to reduce the visual context processed during iterative action denoising. Registers therefore do more than absorb high-norm behavior. They form an effective interface between the world model and the action model, concentrating representations that are directly useful for action generation.

Further analysis shows improved action learning with observation FM loss supervision
even without registers, suggesting that its benefits extend beyond spatial
outlier removal. This motivates treating the current observation not only as
a conditioning input, but also as an auxiliary modeling target. We therefore
introduce noised-observation FM loss supervision, which turns current-scene modeling
into a recovery task from corrupted visual inputs. This objective provides
additional supervision for the world model rather than an additional
action-conditioning stream. RegisterWAM combines this auxiliary objective
with registers and observation FM loss supervision, using registers to condition
action generation.

RegisterWAM achieves 81.64\% overall success on the full LIBERO-Plus test set without using
additional embodied pretraining data or introducing additional foundation models.
On RoboTwin 2.0, it reaches 93.0\% average success, compared with 91.9\%
for FastWAM~\citep{yuan2026fastwam} and 92.6\% for
Faster-WAM~\citep{zhao2026fasterwam}.
Across six real-world task settings, it improves the mean task score
from 0.3483 to 0.5238 over FastWAM.

Our contributions are threefold: \textbf{(i)} We uncover high-norm spatial tokens in WAMs whose action-relevant roles are partially decoupled from their nominal spatial locations. \textbf{(ii)} We propose \emph{RegisterWAM}, which combines non-spatial registers with observation FM loss supervision to separate these roles from spatial tokens and provide an effective interface for action generation. \textbf{(iii)} We introduce auxiliary noised-observation FM loss supervision for current-scene recovery and demonstrate the effectiveness of the resulting framework across robot manipulation benchmarks.

\section{From Spatial Outliers to Explicit Registers}
\label{sec:analysis}

\subsection{Spatial Outliers in World Action Models}
\label{sec:wam_preliminaries}
\label{sec:spatial_outliers}

We consider the FastWAM setting~\citep{yuan2026fastwam}, in which a Video DiT
produces spatial representations that condition an Action DiT. At layer $l$,
we denote the visual hidden states by $H^l=\{h_1^l,\ldots,h_N^l\}$, where
$h_j^l\in\mathbb{R}^d$ corresponds to a location in the latent grid. Our input
resolution yields a $7\times14$ grid and thus $N=98$ spatial tokens. After
normalization and projection, these states form a cached K/V context reused
throughout iterative action denoising. The same tokens therefore serve as both
spatial world representations and conditions for action generation.

\begin{figure*}[t]
    \centering
    {\scriptsize
    \textcolor[HTML]{2E73B7}{\rule{1.8em}{0.7pt}} FastWAM\quad
    \textcolor[HTML]{DF7A18}{\rule{0.7em}{0.7pt}\hspace{0.3em}\rule{0.7em}{0.7pt}} Wan2.2\quad
    \textcolor[HTML]{2A907A}{\rule{0.7em}{0.7pt}\hspace{0.2em}\rule{0.15em}{0.7pt}\hspace{0.2em}\rule{0.5em}{0.7pt}} Cosmos3-Nano}\par
    \begingroup
    \sbox0{\includegraphics{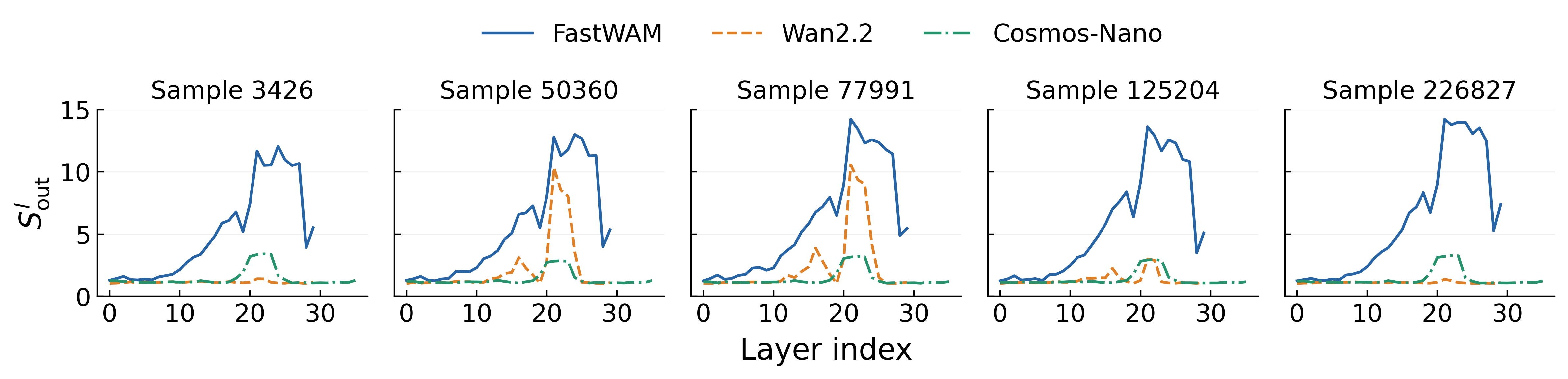}}
    \dimen0=0.16\ht0
    \edef\legendcrop{\the\dimen0}
    \includegraphics[width=\textwidth,trim=0 0 0 \legendcrop,clip]{figures/outliers_analysis/paper_outlier_curves_fastwam_std_5_polished.png}
    \endgroup
    \vspace{-15pt}
    \caption{\small\textbf{Layer-wise spatial outlier strength.} High-norm outliers are more pronounced and consistent in the examined FastWAM checkpoint.}
    \label{fig:outliers_curve_fastwam_and_wan_cosmos}
    \vspace{-0.6em}
\end{figure*}

We measure the largest spatial outlier at layer $l$ by
$S_{\mathrm{out}}^l=\max_j\lVert h_j^l\rVert_2/
\operatorname{median}_j\lVert h_j^l\rVert_2$, i.e., the ratio between the
largest and median spatial-token norms. In FastWAM, this ratio rises sharply in
middle-to-late Video DiT layers and the high-norm locations recur across
observations (Figure~\ref{fig:outliers_curve_fastwam_and_wan_cosmos}). Similar
but weaker and less consistent outliers appear in pretrained Wan2.2~\citep{wan2025} and
Cosmos3-Nano~\citep{nvidia2026cosmos3}. We next test whether the pronounced FastWAM outliers have a
functional consequence for action generation. Visualization settings and
additional examples across models and observations are provided in
Appendix~\ref{app:additional_outliers}.

\subsection{Outliers and Action Generation}
\label{sec:outlier_action}

\begin{figure}[!htbp]
    \centering
    \begin{minipage}[t]{0.242\textwidth}
        \centering
        \includegraphics[width=\linewidth]{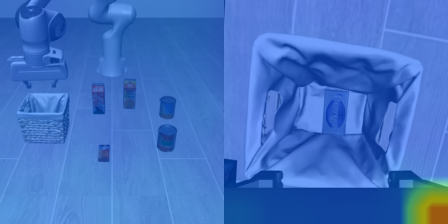}
        \vspace{-0.6em}

        {\small\textbf{(a)} Hidden-state norm}
    \end{minipage}\hfill
    \begin{minipage}[t]{0.242\textwidth}
        \centering
        \includegraphics[width=\linewidth]{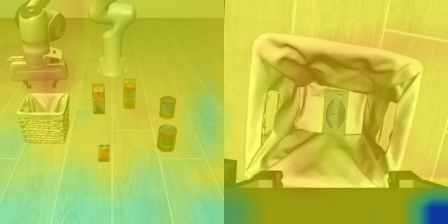}
        \vspace{-0.6em}

        {\small\textbf{(b)} Key norm}
    \end{minipage}\hfill
    \begin{minipage}[t]{0.242\textwidth}
        \centering
        \includegraphics[width=\linewidth]{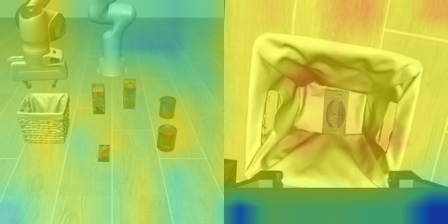}
        \vspace{-0.6em}

        {\small\textbf{(c)} Value norm}
    \end{minipage}\hfill
    \begin{minipage}[t]{0.242\textwidth}
        \centering
        \includegraphics[width=\linewidth]{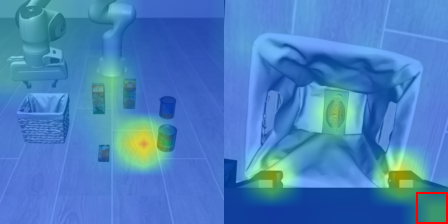}
        \vspace{-0.6em}

        {\small\textbf{(d)} Action sensitivity}
    \end{minipage}
    \vspace{-5pt}
    \caption{\small\textbf{From hidden states to action.} The magnitude anomaly vanishes after K/V projection, while its action relevance persists.}
    \label{fig:outlier_action_relevance}
    \vspace{-0.7em}
\end{figure}

\paragraph{The extreme magnitude does not propagate to the Action DiT.}
A direct explanation for the observed behavior is that the unusually large
hidden norm produces an unusually large action-conditioning signal. The visual
K/V interface largely removes this scale difference:
\begin{equation}
    h_j \longrightarrow \operatorname{Norm}(h_j)
    \longrightarrow (K_j,V_j).
\end{equation}
Although $\lVert h_{\mathrm{out}}\rVert_2 \gg
\lVert h_{\mathrm{normal}}\rVert_2$, the corresponding key and value norms
remain within the range of ordinary spatial tokens. The extreme hidden-state
norm therefore does not directly propagate as an extreme action-facing K/V
magnitude. This rules out the simplest explanation that the outlier affects
action generation merely because its activation scale is unusually large.
Figure~\ref{fig:outlier_action_relevance} compares hidden-state, key, and value
norms with masked-token action sensitivity for the same observation.

\paragraph{Action relevance persists.}
We next directly measure whether each visual token affects action generation.
Let $\mathcal{M}$ denote the default Action DiT attention mask. For visual token $j$, we
construct $\mathcal{M}^{(-j)}$ by masking position $j$ for every Action DiT query, and
define the action sensitivity as the resulting change in the predicted action,
\begin{equation}
    I_j = \left\lVert
    A(K,V;\mathcal{M})-A(K,V;\mathcal{M}^{(-j)})
    \right\rVert_2.
\end{equation}

\begin{wrapfigure}{R}{0.47\textwidth}
    \vspace{-0.6em}
    \centering
    \includegraphics[width=\linewidth]{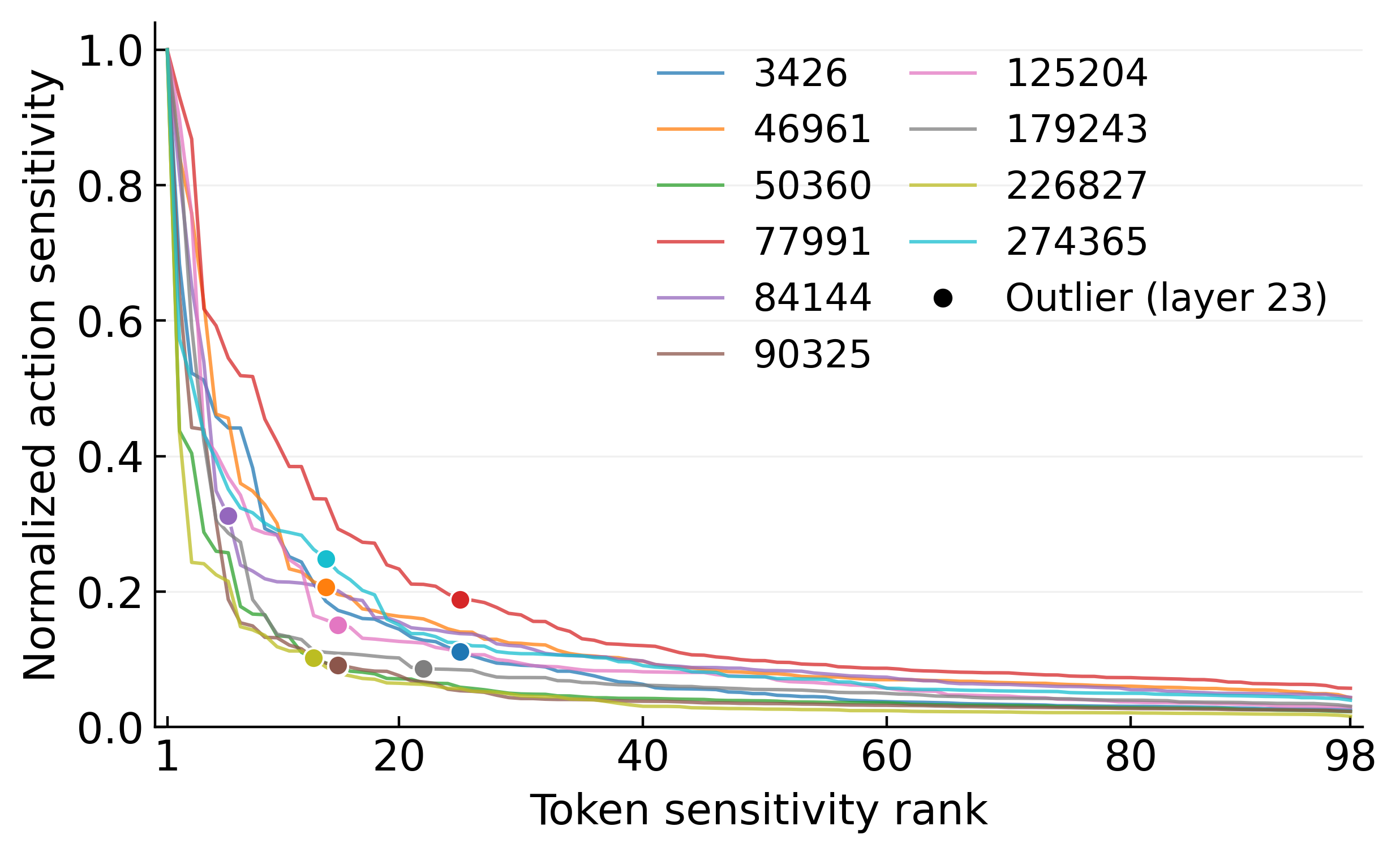}
    \vspace{-0.4em}
    \caption{\small\textbf{Action-sensitivity ranks.} Dots mark the outlier token.}
    \label{fig:outlier_rank_curve}
    \vspace{-0.5em}
\end{wrapfigure}

This intervention removes token $j$ from the action-attention computation,
including its contribution to the attention normalization, while leaving the
remaining visual representations unchanged. The action-sensitivity maps show that the
outlier does not dominate action generation: tokens covering task-relevant
objects can cause substantially larger action changes. Nevertheless, the
outlier exhibits stable, non-trivial action sensitivity across observations. Its rank
is concentrated around
15 among the 98 spatial tokens, rather than behaving like an arbitrary
background token (Figure~\ref{fig:outlier_rank_curve}). Its extreme
hidden-state magnitude is therefore removed at
the interface, yet its downstream action relevance persists.

\paragraph{Conservative interpretation.}
The high-norm outlier should not be viewed as an action token, an attention
sink~\citep{xiao2023attention}, or a unique carrier of action-relevant information. Many ordinary spatial
tokens also contribute substantially to action generation. Instead, the extreme
norm makes one aspect of the model's internal representation behavior directly
observable. The persistence of action relevance after the magnitude anomaly is
removed suggests that spatial token space can support representation behavior
beyond its nominal role of local spatial encoding. The outlier is thus an
observable manifestation of this behavior, rather than the behavior itself.

\subsection{Registers with Observation FM Loss Supervision}
\label{sec:hypothesis_registers}

The preceding analysis suggests that spatial tokens in a WAM can support
representational roles beyond local spatial encoding, with high-norm outliers
providing an observable signature of this behavior. This does not imply that an
outlier uniquely carries action information or identify its semantic content.
Instead, it motivates a controlled architectural probe: \emph{what changes when
the world model is given representation capacity without spatial identity?}

\paragraph{Register tokens.}
Given the spatial video tokens $H$, we append $M$ learnable register tokens
$R=\{r_1,\ldots,r_M\}$ to form the Video DiT input $Z=[H;R]$.
Following the terminology of \citet{darcet2024vitregisters}, we call these
learnable tokens \emph{registers}. The spatial tokens and registers are jointly
updated by the same self-attention layers, while registers have no explicit
correspondence to image locations. We use the spatial outlier as a probe of
whether this additional non-spatial capacity changes the observable
organization of the representation.

\paragraph{Registers alone are insufficient.}
Registers alone do not fully eliminate spatial outliers: outliers
remain at the output layer, even though high-norm responses also emerge among
the registers (Figure~\ref{fig:redistribution}a). Layer-wise curves in
Appendix~\ref{app:additional_outliers} show that intermediate-layer peaks are
nevertheless suppressed. The limited output-layer change is accompanied by
a modest policy gain in the naive setting. FastWAM conditions its Action DiT
on the dense observation states (\emph{Raw}, referring only to observation
tokens in this section), $A\leftarrow H$, and achieves 53.64\% success. Appending
8 registers and conditioning on both token types, $A\leftarrow[H;R]$,
without observation FM loss supervision reaches 54.18\%. Thus, simply adding explicit
slots neither resolves the spatial anomaly nor substantially improves the policy.

\paragraph{Observation FM loss supervision reorganizes the representation.}
The remaining output-layer outliers suggest that extra non-spatial capacity
does not by itself constrain how the original spatial tokens are used. We
therefore introduce observation FM loss supervision on the dense clean
observation tokens. Specifically, we apply the standard flow-matching loss at
zero noise level ($\sigma=0$) to the dense observation-token outputs, excluding
registers. Under this supervision, the output-layer
spatial-token norms become substantially more uniform: the pronounced spatial
outlier is reduced, while high-norm responses remain concentrated in the
registers (Figure~\ref{fig:redistribution}b). We do not interpret this
redistribution as evidence that any individual register has a fixed semantic
role; rather, it shows that the observable specialization no longer needs to
occupy a spatial location.
Figure~\ref{fig:reg_fm_outlier_curves} in Appendix~\ref{app:additional_outliers}
further shows their complementary effects across layers: registers suppress
intermediate-layer outliers, observation FM loss supervision primarily suppresses
final-layer outliers, and their combination suppresses both.

\begin{figure*}[t]
\centering
\resizebox{0.96\linewidth}{!}{%
\begin{tabular}{@{}r@{\hspace{0.65em}}c@{\hspace{0.35em}}c@{\hspace{0.65em}}c@{}}
    \raisebox{0.30in}{\small (a)} &
    \includegraphics[height=0.69in]{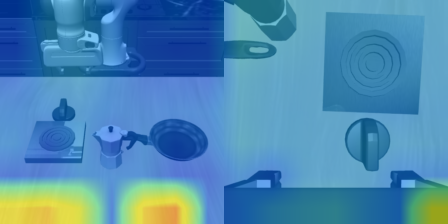} &
    \includegraphics[height=0.69in]{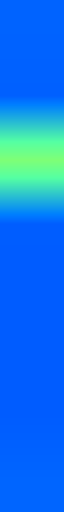} &
    \includegraphics[height=0.75in,trim=0 12 0 10,clip]{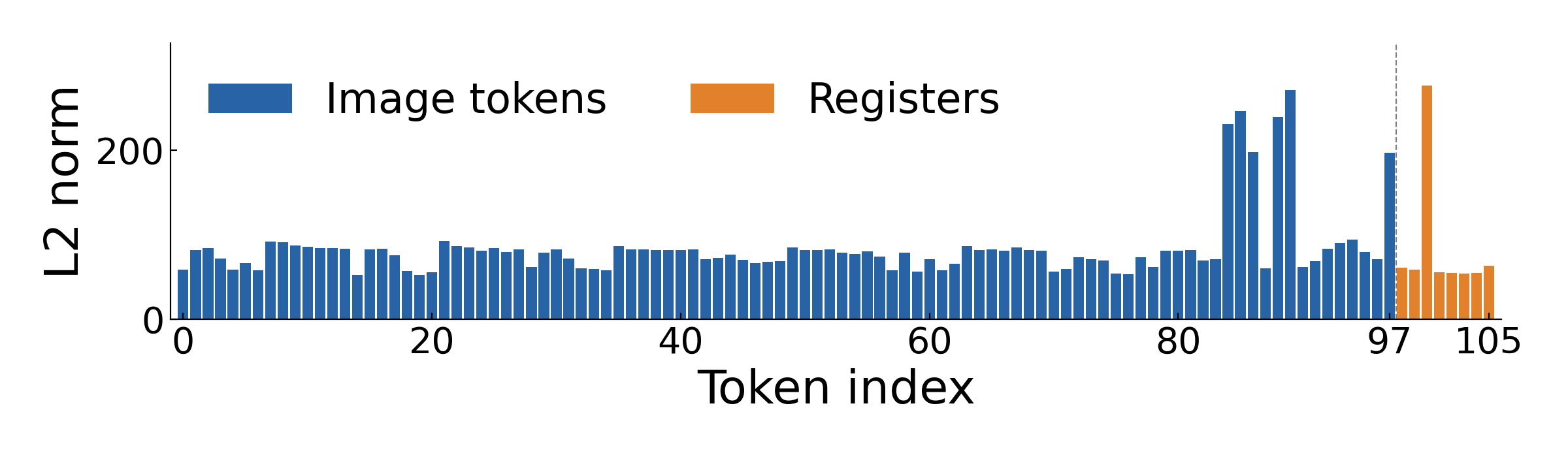}\\[0.65em]
    \raisebox{0.30in}{\small (b)} &
    \includegraphics[height=0.69in]{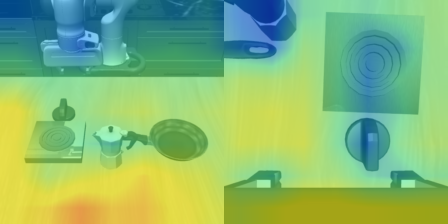} &
    \includegraphics[height=0.69in]{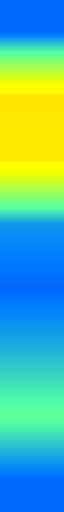} &
    \includegraphics[height=0.75in,trim=0 12 0 10,clip]{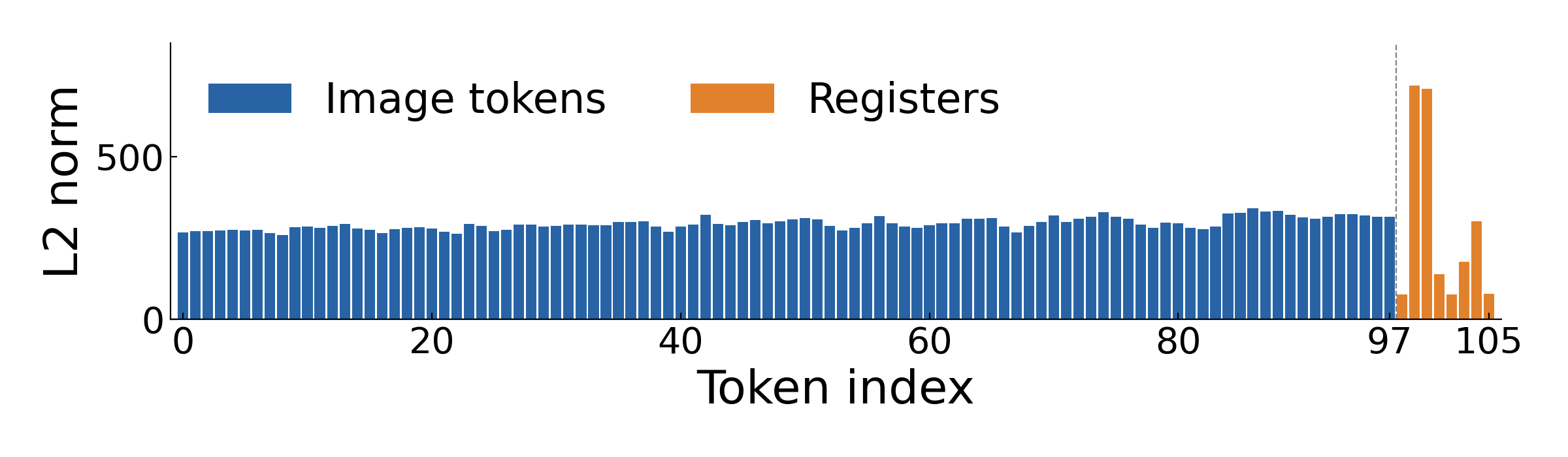}
\end{tabular}%
}
\caption{\small\textbf{Output-layer representations with registers.} (a) Without observation FM loss supervision, spatial outliers remain alongside high-norm registers. (b) With observation FM loss supervision, spatial-token norms are regularized while high-norm responses remain concentrated in registers.}
\label{fig:redistribution}
\end{figure*}

\paragraph{Action conditioning after outlier reorganization.}
\begin{wrapfigure}{R}{0.39\textwidth}
    \centering
    \vspace{-0.8em}
    \includegraphics[width=\linewidth]{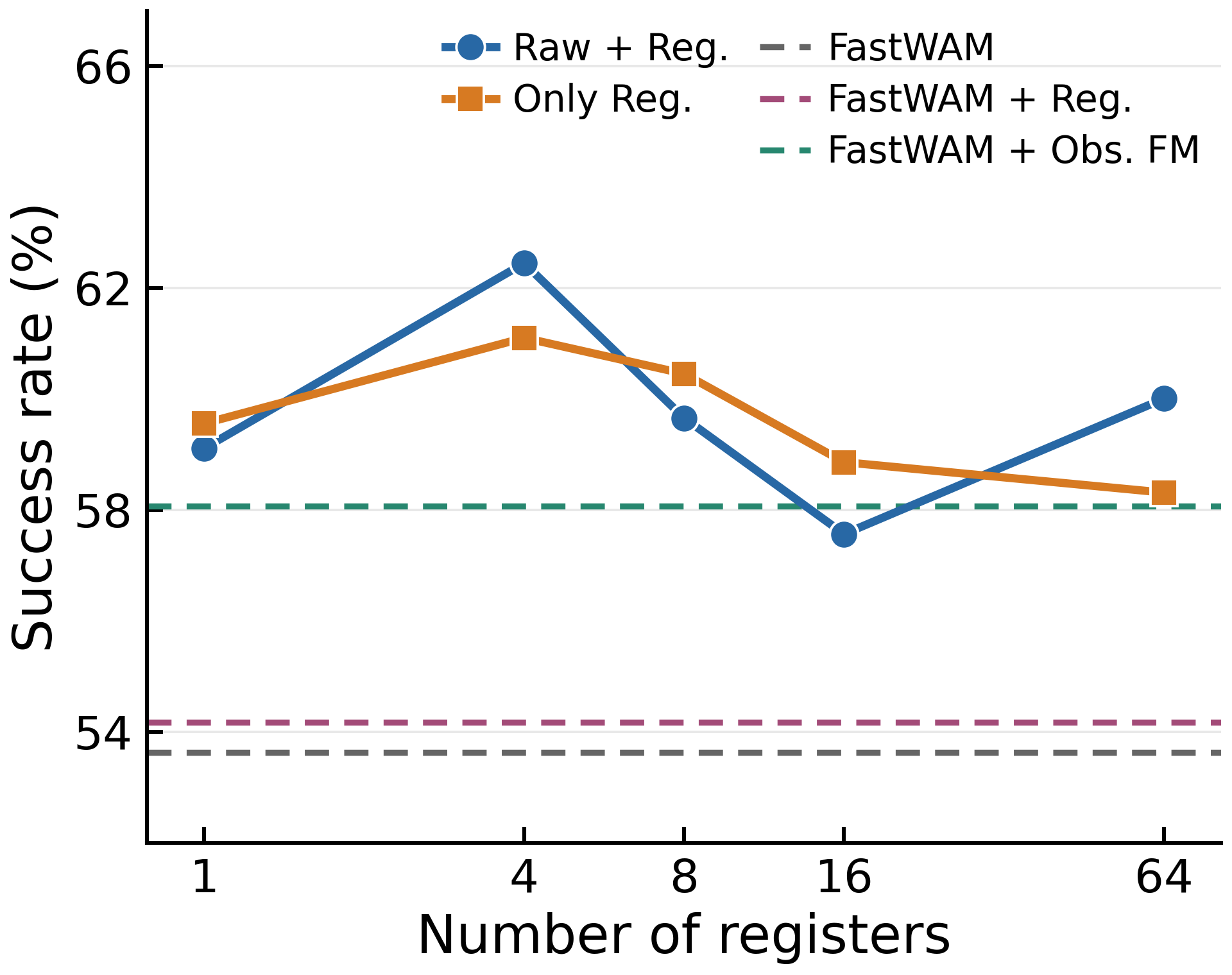}
    \vspace{-0.7em}
    \caption{\small\textbf{Register interfaces.}
    Curves compare Only Reg. and Raw + Reg. conditioning with
    observation FM loss supervision. Raw denotes dense observation tokens only.
    Dashed lines show FastWAM, raw plus 8 registers without observation FM
    loss supervision, and raw tokens with observation FM loss supervision but
    no registers.}
    \label{fig:register_count_analysis}
    \vspace{-0.8em}
\end{wrapfigure}

After observation FM loss supervision suppresses the output-layer spatial outlier, we ask
how the reorganized world states should condition action generation. We compare
two interfaces while varying the number of registers. The first exposes only
the register states to the Action DiT, $A\leftarrow R$, directly testing
whether the registers contain sufficient action-relevant
information. With 1, 4, 8, 16, and 64 registers, this interface obtains
59.56\%, 61.11\%, 60.46\%, 58.86\%, and 58.32\%, respectively. Every displayed
count remains competitive with the 58.07\% raw-token reference trained with observation FM loss supervision, and 4 registers give the strongest register-only result among
the displayed settings. Increasing the number of registers does not consistently
improve performance.

The second interface exposes both the raw observation states and registers,
$A\leftarrow[H;R]$. Under the same observation FM loss supervision, the
corresponding results are
59.11\%, 62.45\%, 59.66\%, 57.57\%, and 60.01\%. Raw tokens improve upon the
register-only interface at 4 and 64 registers, but not at 1, 8, or 16. They can
therefore complement registers in some configurations, although the combined
interface is not uniformly superior.

Together, the representation and policy results form a consistent sequence.
Registers alone provide a possible destination for non-spatial specialization,
but do not fully eliminate spatial outliers or substantially improve the
policy in the naive setting. Observation FM loss supervision reorganizes the token norms; after this reorganization,
the resulting register states become directly useful as action conditions. A
small register set can carry action-relevant information on its own, while raw
spatial tokens remain complementary in some configurations. We therefore treat
registers as explicit representation slots, not as a universal replacement for
dense visual conditioning. Given their broadly comparable performance in
Figure~\ref{fig:register_count_analysis}, we adopt register-only conditioning
to reduce the visual context processed during action denoising.

\subsection{Beyond Outlier Reorganization}
\label{sec:observation_supervision}

The preceding results resolve the spatial-outlier observation: registers
provide non-spatial slots, and observation FM loss supervision causes the visible
specialization to move away from spatial locations. A separate control reveals
an additional effect of the same supervision. Without introducing any
registers, the setting trained with observation FM loss supervision achieves
58.07\% (the green dashed reference in Figure~\ref{fig:register_count_analysis}),
compared with 53.64\% for the original FastWAM.
This improvement does not require redistribution into registers.

This result motivates treating the current observation not only as a
conditioning input, but also as an auxiliary modeling target. We extend this
idea by perturbing the observation with noise and introducing
noised-observation FM loss supervision. The resulting task requires predicting a recovery direction from
incomplete visual information, making current-scene modeling an explicit
denoising task. Its role is to provide additional supervision for the world
model, rather than additional representations for the action model to read.
The next section combines noised-observation FM loss supervision with registers
and observation FM loss supervision to construct RegisterWAM.

\section{RegisterWAM}
\label{sec:method}

\begin{figure}[htbp]
\centering
\includegraphics[width=0.98\linewidth]{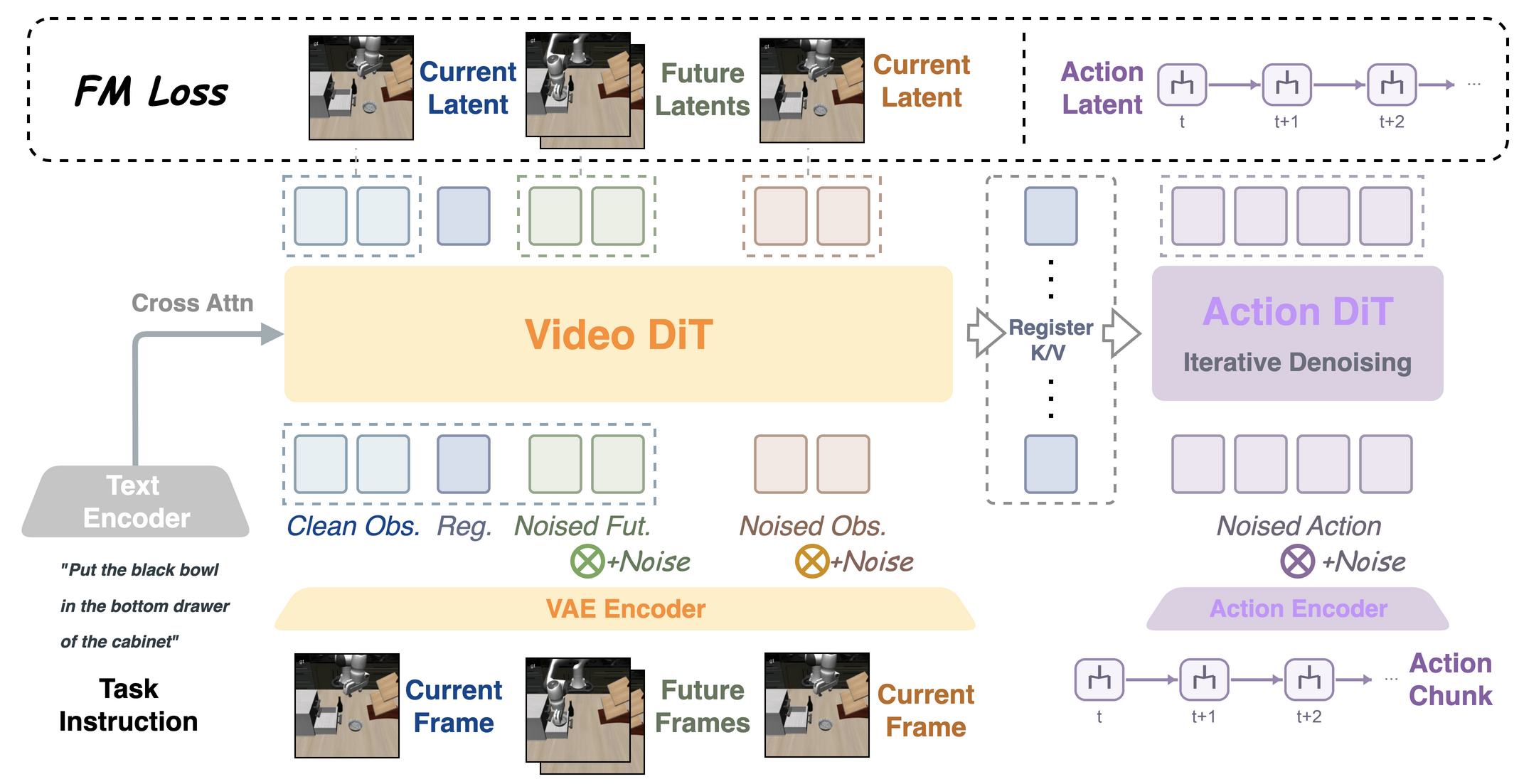}
\caption{\small\textbf{RegisterWAM architecture.}
Registers condition the Action DiT; visual outputs receive FM loss supervision.
Encoded robot-state tokens are appended to the text embeddings (omitted for clarity).}
\label{fig:registerwam_architecture}
\vspace{-0.6em}
\end{figure}

Building on Section~\ref{sec:analysis}, RegisterWAM combines learnable
registers with observation FM loss supervision.
Registers provide the visual conditioning for the Action DiT.
Noised-observation FM loss supervision supplies an auxiliary denoising task. Figures~\ref{fig:registerwam_architecture}
and~\ref{fig:training_attention_mask} show the architecture and attention mask.

\subsection{Registers as the World--Action Interface}
\label{sec:register_conditioning}

\paragraph{World representation.}
We encode the current observation into VAE latents and append $M$ learnable
register tokens to the visual sequence. The Video DiT processes these tokens
together with future tokens, whose inputs are pure Gaussian noise during
both training and inference. During training, encoded future frames provide
the targets for the future FM loss. Language instructions are encoded
into text embeddings, followed by encoded robot-state tokens; the resulting
sequence supplies cross-attention conditioning.

\paragraph{Action conditioning.}
Let $R^{\ell}$ denote the register states at Video DiT layer $\ell$.
We construct the corresponding action-conditioning keys and values from
these register states, following the layer-wise K/V conditioning of the
base WAM. This register K/V cache supplies the visual conditioning for the Action DiT.
Thus, the world model retains dense visual representations while the
visual interface to the action model consists of $M$ registers.

\subsection{Observation-Side FM Loss Supervision}
\label{sec:observation_objectives}

\paragraph{Observation FM loss supervision.}
Motivated by the outlier analysis, we apply the observation FM loss
$\mathcal{L}_{\mathrm{obs}}$ to the outputs corresponding to the dense clean
observation tokens. Together with registers, this supervision reduces
spatial outliers and concentrates high-norm responses in registers
(Section~\ref{sec:analysis}). Registers themselves are not supervised by
visual FM losses. The improvement with observation FM loss supervision
even without registers further motivates extending observation FM loss supervision
to perturbed inputs.

\begin{wrapfigure}{R}{0.43\textwidth}
\centering
\vspace{-0.5em}
\includegraphics[width=\linewidth]{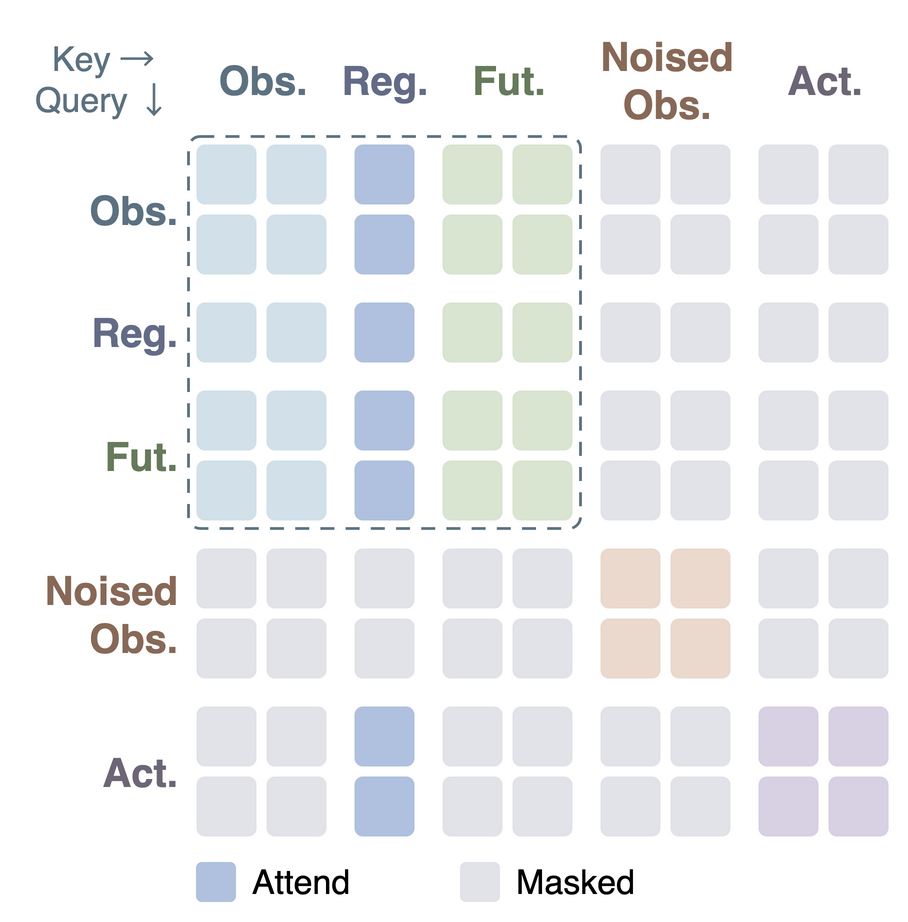}
\caption{\small\textbf{Training attention mask.}
Action tokens attend to registers and themselves. Gray entries are masked.}
\label{fig:training_attention_mask}
\vspace{-0.3em}
\end{wrapfigure}

\paragraph{Noised-observation FM loss supervision.}
Given the current observation latent $x$, we construct
\begin{equation}
 x_\sigma=(1-\sigma)x+\sigma\epsilon,
 \qquad \epsilon\sim\mathcal{N}(0,I),
\end{equation}
where $\sigma$ controls the noise level. The corresponding tokens are
processed by the same Video DiT, with the noised-observation FM loss
$\mathcal{L}_{\mathrm{noised\text{-}obs}}$ on their outputs.
The attention mask in Figure~\ref{fig:training_attention_mask} isolates
this branch, which provides an auxiliary denoising
task whose gradients update the Video DiT parameters, not an additional
visual conditioning stream.

\subsection{Training and Inference}

We jointly train the Video DiT and Action DiT with
\begin{equation}
\mathcal{L}=\lambda_a\mathcal{L}_{\mathrm{action}}
+\lambda_v\bigl(\mathcal{L}_{\mathrm{obs}}+\mathcal{L}_{\mathrm{future}}\bigr)
+\lambda_n\mathcal{L}_{\mathrm{noised\text{-}obs}},
\end{equation}
where $\mathcal{L}_{\mathrm{action}}$ supervises action-chunk denoising and
the other three terms supervise the corresponding dense visual outputs.
All visual losses exclude register tokens. Future inputs remain at the
maximum noise level during training to match their inference-time inputs,
while noised-observation FM loss supervision uses a separate noise level $\sigma$.
At inference, the Video DiT runs once to construct the register K/V cache,
which the Action DiT reuses throughout its iterative action denoising.
The action model therefore accesses a fixed-size register interface without
requiring iterative video generation.

\section{Experiments}
\label{sec:experiments}

Section~\ref{sec:analysis} establishes the FastWAM-base representation findings.
Here we evaluate the resulting model, compare action-conditioning choices,
ablate observation-side FM loss supervision, and test transfer to another generative backbone.

\subsection{Experimental Setup}
We evaluate on LIBERO-Plus~\citep{fei25libero-plus}, RoboTwin 2.0~\citep{chen2025robotwin},
and 6 real-world task settings.
The world model builds on Wan2.2, while the policy uses an Action DiT with 10 denoising steps. 
Unless noted otherwise, RegisterWAM uses 8 registers interacting with observation and future tokens, and noise level $\sigma=0.5$ for the noised-observation FM loss supervision. 
Comparisons include published FastWAM and other WAMs, representative
vision--language--action policies, and our future-conditioned baseline. 
For our LIBERO-Plus experiments, all models are trained exclusively
on LIBERO~\citep{liu2023libero} data, without using LIBERO-Plus data for training.
The ablation studies are evaluated on the same 20\% subset of the
LIBERO-Plus test data.
All models with registers use an action-loss weight $\lambda_a=0.2$.
For the FastWAM analysis in Section~\ref{sec:analysis}, the 58.07\% no-register
reference with observation FM loss supervision also uses $\lambda_a=0.2$,
whereas the original 53.64\% FastWAM uses $\lambda_a=1.0$.
Changing only the action-loss weight from $1.0$ to $0.2$ leaves the original
FastWAM's performance essentially unchanged.
All settings in the final-model ablations use $\lambda_a=0.2$.

\subsection{Main Results}
\begin{table*}[t]
\centering
\small
\caption{\small LIBERO-Plus success rates (\%). Emb. PT.: embodied-policy pretraining; Extra Prior: external pretrained models, modalities, or geometric/semantic/motion supervision.
}
\label{tab:libero_plus}
\setlength{\tabcolsep}{2.4pt}
\renewcommand{\arraystretch}{0.88}
\resizebox{0.90\textwidth}{!}{%
\begin{tabular}{@{}lccrrrrrrrr@{}}
\toprule
Method & \shortstack{Emb.\\PT.} & \shortstack{Extra\\Prior} & Camera & Robot & Lang. & Light & Backg. & Noise & Layout & Avg. \\
\midrule
$\pi_{0.5}$~\cite{intelligence2025pi_} & $\checkmark$ & -- & 75.4 & 77.5 & 85.6 & 96.9 & 94.6 & 89.7 & 85.7 & 85.7 \\
FastWAM~\cite{yuan2026fastwam} & -- & -- & 16.4 & 44.5 & 68.9 & 78.2 & 53.7 & 37.7 & 60.7 & 51.5 \\
Faster-WAM~\cite{ma2026fasterwam} & -- & -- & 67.9 & 49.0 & 92.1 & 94.3 & 57.0 & 82.3 & 82.7 & 75.0 \\
Faster-WAM~\cite{zhao2026fasterwam} & -- & -- & 53.8 & 71.6 & 94.7 & 96.3 & 61.3 & 63.6 & 79.1 & 73.6 \\
\midrule
4D-WAM~\cite{yang20264dwam} & -- & $\checkmark$ & 45.2 & 64.3 & 90.6 & 94.3 & 57.7 & 69.1 & 79.2 & 71.0 \\
ST-WAM~\cite{wang2026stwam} & -- & $\checkmark$ & 55.4 & 60.1 & 79.3 & 93.0 & 74.2 & 79.5 & 74.3 & 72.8 \\
DreamWAM~\cite{yuan2026dreamwam} & -- & $\checkmark$ & 53.8 & 63.6 & 94.8 & 96.7 & 71.6 & 67.2 & 80.7 & 75.5 \\
GaussianWAM~\cite{zhang2026gaussianwam} & -- & $\checkmark$ & 54.7 & 61.6 & 75.4 & 90.8 & 58.2 & 83.0 & 76.7 & 71.3 \\
Robust-WAM~\cite{yan2026robustwam} & -- & $\checkmark$ & 28.7 & 43.4 & 73.6 & 90.2 & 62.6 & 56.8 & 67.8 & 58.9 \\
PILOT~\cite{ma2026pilot} & -- & $\checkmark$ & 76.0 & 70.0 & 84.0 & 92.0 & 85.0 & 78.0 & 82.0 & 81.0 \\
\midrule
\textbf{RegisterWAM} & -- & -- & \textbf{59.2} & \textbf{71.9} & \textbf{93.7} & \textbf{95.7} & \textbf{87.9} & \textbf{85.1} & \textbf{84.3} & \textbf{81.6} \\
\bottomrule
\end{tabular}
}
\vspace{14pt}
\end{table*}

Table~\ref{tab:libero_plus} reports performance across the seven LIBERO-Plus
distribution shifts. RegisterWAM achieves 81.64\% overall success on the full
LIBERO-Plus test set of 10,030 episodes,
exceeding the listed FastWAM and Faster-WAM results while using
the same Wan2.2 backbone, without embodied-policy pretraining or additional
geometric or semantic teacher models. We separately mark methods that use
additional pretrained models, modalities, or externally supplied geometric,
semantic, or motion supervision, excluding FM losses on the original visual data,
since their reported scores reflect a different resource setting. Its highest category scores are under language, lighting, and background changes.
The same model achieves 82.52\% on the 20\% test subset used for the
ablation studies.

\begin{table}[t]
\centering
\begin{minipage}[t]{0.60\linewidth}
\centering\small
\caption{\small RoboTwin 2.0 success rates (\%).}
\label{tab:robotwin}
\setlength{\tabcolsep}{3.2pt}
\renewcommand{\arraystretch}{1.03}
\begin{tabular}{@{}lcccc@{}}
\toprule
Method & \mbox{Emb. PT.} & Clean & Rand. & Avg. \\
\midrule
LingBot-VA~\cite{lingbot-va2026} & $\checkmark$ & 92.9 & 91.5 & 92.2 \\
FastWAM~\cite{yuan2026fastwam} & -- & 91.9 & 91.8 & 91.9 \\
Faster-WAM~\cite{zhao2026fasterwam} & -- & 92.8 & 92.3 & 92.6 \\
\midrule
\textbf{RegisterWAM} & -- & \textbf{92.9} & \textbf{93.0} & \textbf{93.0} \\
\bottomrule
\end{tabular}
\end{minipage}\hfill
\begin{minipage}[t]{0.34\linewidth}
\centering\small
\caption{\small ImageWAM transfer (\%).}
\label{tab:imagewam_transfer}
\setlength{\tabcolsep}{5pt}
\begin{tabular}{@{}lr@{}}
\toprule
Method & Avg. \\
\midrule
\shortstack[l]{ImageWAM\\{\scriptsize (FLUX.2 4B)}} & 83.10 \\[3pt]
\shortstack[l]{ImageWAM + Ours\\{\scriptsize (FLUX.2 4B)}} & \textbf{85.42} \\
\bottomrule
\end{tabular}

\end{minipage}
\vspace{14pt}
\end{table}

On RoboTwin 2.0 (Table~\ref{tab:robotwin}), RegisterWAM achieves 92.9\% in the
clean setting and 93.0\% under randomization, for a 93.0\% average. This
slightly exceeds the compared FastWAM and Faster-WAM variants without using
embodied-policy pretraining.

\begin{table}[t]\centering\small
\caption{\small Real-world task scores (20 trials per setting). OOD variants modify the bottle-handover task; Avg. equally weights all six settings.}\label{tab:real}
\setlength{\tabcolsep}{3pt}
\begin{tabular}{@{}lrrrrrrr@{}}
\toprule
& \multicolumn{3}{c}{In-distribution} & \multicolumn{3}{c}{Out-of-distribution} & \\
\cmidrule(lr){2-4}\cmidrule(lr){5-7}
Method & Batteries & Stacking & Handover & Height & Tablecloth & Distractor & Avg. \\
\midrule
FastWAM & 0.5000 & 0.0000 & 0.6200 & 0.5500 & \textbf{0.0000} & 0.4200 & 0.3483 \\
RegisterWAM & \textbf{0.6875} & \textbf{0.2500} & \textbf{0.7550} & \textbf{0.7300} & \textbf{0.0000} & \textbf{0.7200} & \textbf{0.5238} \\
\bottomrule
\end{tabular}
\vspace{14pt}
\end{table}

\begin{figure}[htbp]
\centering
\begin{minipage}[t]{0.32\linewidth}
\centering
\includegraphics[width=\linewidth]{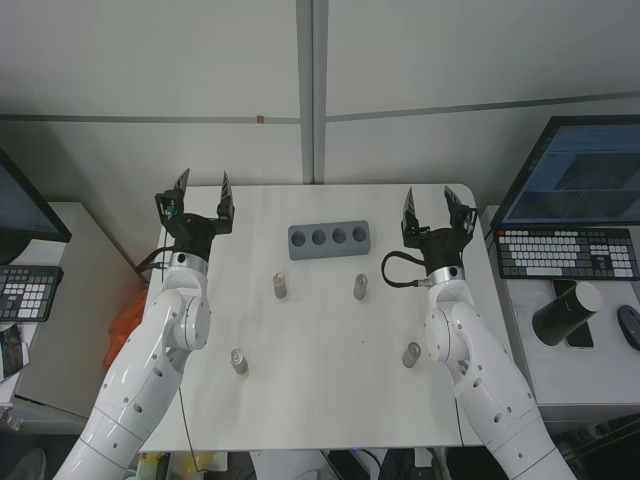}
{\small Battery insertion}
\end{minipage}\hfill
\begin{minipage}[t]{0.32\linewidth}
\centering
\includegraphics[width=\linewidth]{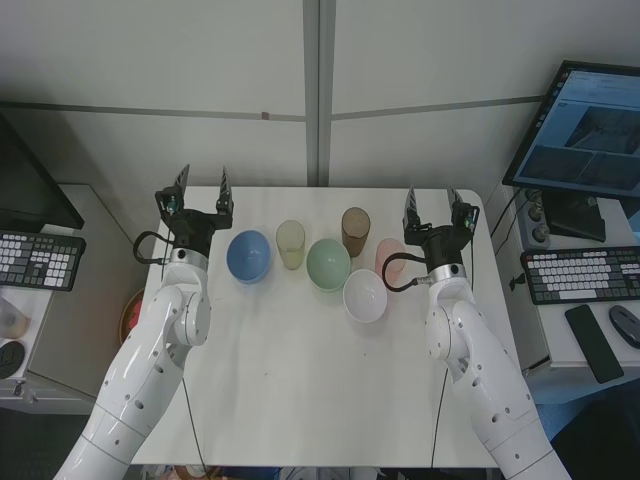}
{\small Bowl-and-cup stacking}
\end{minipage}\hfill
\begin{minipage}[t]{0.32\linewidth}
\centering
\includegraphics[width=\linewidth]{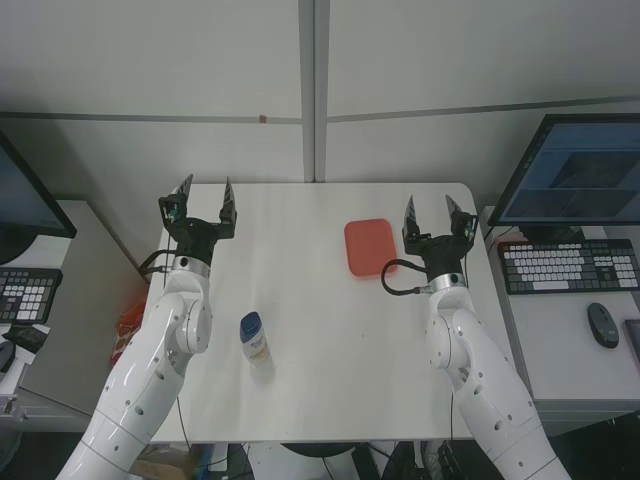}
{\small Bottle handover}
\end{minipage}
\caption{\small Scenes for the three real-world ID tasks. The three OOD variants modify the bottle-handover setting.}
\label{fig:real_world_tasks}
\vspace{-0.6em}
\end{figure}

\paragraph{Real-world evaluation.}
We compare RegisterWAM with FastWAM on a three-camera Tianji dual-arm platform
on 3 in-distribution (ID) tasks: battery insertion, bowl-and-cup stacking, and
bottle handover and placement (Figure~\ref{fig:real_world_tasks}), using
300, 60, and 300 demonstrations, respectively.
For battery insertion, the left arm inserts the two left batteries into the
two left slots, followed by the right arm inserting the two right batteries
into the two right slots. The stacking task forms a stack of three bowls
and then a stack of three cups. For bottle handover, the left arm grasps and
lifts a white shampoo bottle, passes it to the right arm, and the right arm
places it into the orange plate. All 3 out-of-distribution (OOD) settings
modify this last task: (i) changing the table height while keeping the arm
height fixed; (ii) covering the original white table with a blue tablecloth;
and (iii) adding a white distractor plate beside the target plate.

We evaluate each of the 6 settings over 20 trials. Using step-wise task scores rather than binary
success rates, RegisterWAM improves the mean score from 0.3483 to 0.5238
(Table~\ref{tab:real}). It improves five settings, while both methods score
zero under the blue-tablecloth shift. Additional evaluation and scoring details,
along with the full task instructions, are provided in Appendix~\ref{app:real_world}.

\subsection{Ablation Studies}
\label{sec:ablation}
\paragraph{Establishing a strong baseline.}
Following future conditioning in Faster-WAM~\citep{zhao2026fasterwam}
and RIFT~\citep{zhang2026rift}, our future-conditioned baseline conditions actions on dense
observation and future tokens (\emph{Raw}). We use pure-noise future inputs
during both training and inference throughout the ablations. This future-conditioned baseline
achieves 76.29\% success on the LIBERO-Plus subset.

\begin{table}[t]
\centering
\renewcommand{\arraystretch}{1}
\begin{minipage}[t]{0.44\linewidth}
\centering\small
\caption{\small Cumulative component ablation.}
\label{tab:conditioning}
\setlength{\tabcolsep}{3pt}
\begin{tabular}[t]{@{}lcr@{}}
\toprule
Configuration & Interface & Suc. (\%) \\
\midrule
Baseline & Raw & 76.29 \\
+ Noised Obs. FM & Raw & 78.59 \\
+ Reg. and Obs. FM & Reg. & \textbf{82.52} \\
\bottomrule
\end{tabular}

\end{minipage}\hfill
\begin{minipage}[t]{0.23\linewidth}
\centering
\small
\caption{\small\mbox{Register count.}}
\label{tab:register_count}
\begin{tabular}[t]{@{}cr@{}}
\toprule
Registers & Suc. (\%) \\
\midrule
4 & 78.49 \\
8 & \textbf{82.52} \\
16 & 81.03 \\
\bottomrule
\end{tabular}

\end{minipage}\hfill
\begin{minipage}[t]{0.23\linewidth}
\centering\small
\caption{\small Noise level.}
\label{tab:noised}
\begin{tabular}[t]{@{}cr@{}}
\toprule
Nominal $\sigma$ & Suc. (\%) \\
\midrule
0.25 & 76.14 \\
0.50 & \textbf{82.52} \\
0.75 & 80.98 \\
\bottomrule
\end{tabular}

\end{minipage}
\vspace{14pt}
\end{table}

\paragraph{FM loss supervision and the register interface.}
Adding isolated noised-observation FM loss supervision raises success from 76.29\%
to 78.59\% while keeping the action conditioning unchanged
(Table~\ref{tab:conditioning}). Replacing the raw interface with 8 registers
and adding observation FM loss supervision further improves success to 82.52\%.

\paragraph{Key parameters.}
We vary the register count and noised-observation noise level separately in
the full model. Among the tested register counts, 8 registers perform best;
increasing the count to 16 does not improve performance
(Table~\ref{tab:register_count}). For the noise-level sweep, the scheduler's
timestep-dependent loss weighting means that varying $\sigma$ changes both
input corruption and the effective auxiliary FM loss weight. Performance
peaks at $\sigma=0.5$ and drops markedly at the lower noise level of
$\sigma=0.25$ (Table~\ref{tab:noised}). We therefore use 8 registers and
$\sigma=0.5$ by default.

\paragraph{Transfer to ImageWAM.}
We test transfer to ImageWAM~\citep{zhang2026imagewam} because it uses the
FLUX.2 image-editing backbone~\citep{flux-2-2025} rather than the video-generation backbone used here.
Applying the same design improves
average success on LIBERO-Plus from 83.10\% to 85.42\%, a gain of
2.32 percentage points over the original model
(Table~\ref{tab:imagewam_transfer}).

Additional register-visualization analysis is provided in
Appendix~\ref{app:register_visualization}.
Related work is discussed in Section~\ref{sec:related}.

\section{Related Work}
\label{sec:related}

\paragraph{World Action Models.}
WAMs connect visual prediction to robot control through generated visual plans
\citep{du2023learninguniversalpoliciestextguided,bharadhwaj2024gen2act,zhou2024robodreamer,feng2025vidarembodiedvideodiffusion},
predictive features~\citep{hu2024video,ge2025,pai2025mimicvideo,ma2026dit4dit},
or joint video--action learning
\citep{cheang2024gr2generativevideolanguageactionmodel,liang2025videogenerators,li2025unified,zhu2025unifiedworldmodelscoupling}.
Other video--action architectures extend this direction
\citep{kim2026cosmos,bi2025motusunifiedlatentaction,lingbot-va2026,ye2026world,motubrain2026}.
Related VLA approaches predict visual goals or structured world knowledge
before action generation~\citep{zhao2025cotvlavisualchainofthoughtreasoning,dreamvla25}.
Recent designs reduce inference cost through cached representations and
lighter action modules
\citep{yuan2026fastwam,ma2026fasterwam,zhao2026fasterwam,zhang2026rift},
action-centered causal modeling~\citep{ye2026gigaworld},
compact backbones~\citep{li2026light,li2026efficientwam}, or image editing
instead of video prediction~\citep{zhang2026imagewam}.
Others add geometric, semantic, motion, or multi-stream supervision
\citep{yang20264dwam,wang2026stwam,yuan2026dreamwam,zhang2026gaussianwam,yan2026robustwam,yan2026flexpi,ma2026pilot}.
Their common premise is that visual modeling supplies useful representations
or learning signals for action prediction. We complement these efforts by
studying token-level representation organization and the effects of registers
and observation FM loss supervision on spatial outliers and action learning.

\paragraph{Register tokens and latent action interfaces.}
Registers accommodate non-spatial computation in ViTs
\citep{darcet2024vitregisters} and also benefit pixel-space DiTs without
naturally occurring patch-token outliers~\citep{starodubcev2026registers}.
High-norm outliers differ from attention sinks, which concern attention
mass~\citep{xiao2023attention}.
Driving on Registers~\citep{kirby2026driving} and
Latent Interface Training~\citep{lin2026latent} explore compact latent
interfaces for control.
Our focus is not latent-token conditioning itself, but spatial outliers and
their persistent action relevance in WAMs: we study how observation FM loss
supervision reorganizes high-norm responses between dense spatial tokens and
registers, and use registers as the world--action interface.

\section{Conclusion}
We identified persistent high-norm spatial outliers in the examined WAM and
found that their action relevance survives the removal of the magnitude
anomaly at the visual K/V interface. Registers alone do not fully eliminate spatial outliers.
Observation FM loss supervision of dense observation tokens suppresses
output-layer outliers while high-norm responses
concentrate in registers. The resulting register states support effective
action conditioning. Improved action performance is also observed with observation FM loss
supervision without registers, motivating additional noised-observation
FM loss supervision. These findings connect explicit non-spatial capacity and
observation-side FM loss supervision as two components of WAM design.

\bibliography{references}
\bibliographystyle{plainnat}

\newpage
\appendix
\section{Additional Outlier Analysis}
\label{app:additional_outliers}

\paragraph{Visualization settings.}
For each Video DiT layer, we compute the $\ell_2$ norm independently for every
spatial token and reshape the resulting values to the latent spatial grid. The
heatmaps in the main paper and below are resized to the input-image resolution
only for visualization and shown alone or overlaid on the corresponding observation; no
spatial smoothing is used for the reported token statistics. The main-paper
visualizations use representations from layers 21--23.
Hidden-state, key, and value maps show
$\lVert h_j\rVert_2$, $\lVert K_j\rVert_2$, and $\lVert V_j\rVert_2$,
respectively. Action-sensitivity maps show $I_j$ from
Section~\ref{sec:outlier_action}, obtained by masking one visual position from
all Action DiT queries at a time. Colors should be compared spatially within
each panel rather than interpreted as a common absolute scale across images;
cross-model magnitude comparisons use the normalized statistic $S_{\mathrm{out}}$.
Unless otherwise specified, the main-paper outlier visualizations use future
noise level $\sigma_{\mathrm{future}}=1$, i.e., pure-noise future inputs.

\begin{figure}[H]
    \centering
    \includegraphics[width=0.74\linewidth]{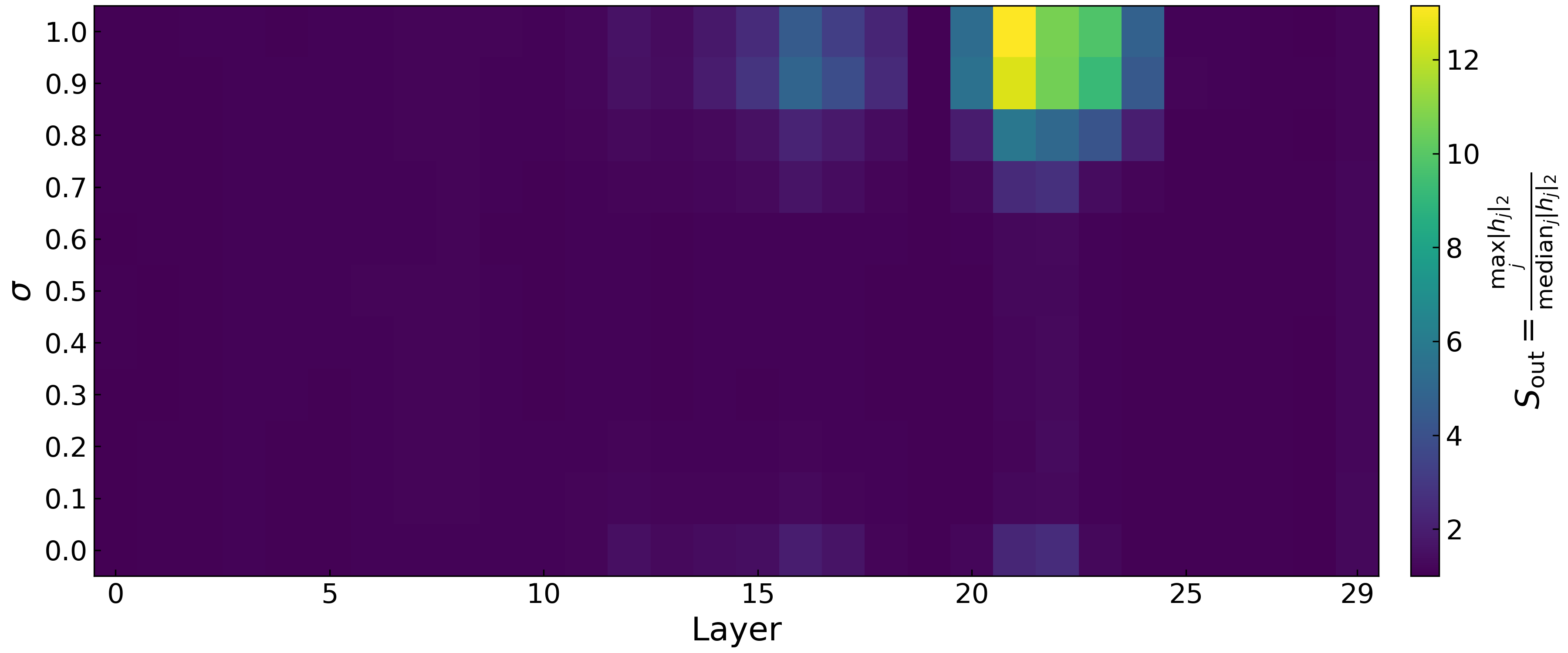}
    \caption{\small\textbf{Wan2.2 first-frame outliers versus future noise.}
    Color denotes first-frame spatial outlier strength $S_{\mathrm{out}}$.
    Pronounced peaks occur mainly in middle-to-late layers under strong future noise.}
    \label{fig:wan_future_noise_outliers}
    \end{figure}

\paragraph{First-frame outliers under different future noise levels.}
Figure~\ref{fig:wan_future_noise_outliers} examines the pretrained Wan2.2
backbone while varying the noise level of its future-frame inputs.
The horizontal axis indexes Video DiT layers, the vertical axis gives the
future noise level, and color shows $S_{\mathrm{out}}$ computed over spatial
tokens of the first frame. Thus, the varied noise belongs to the future
inputs, whereas the measured outliers belong to the first-frame representation.
This sweep is distinct from the noised-observation objective in
Section~\ref{sec:method}.

The strongest first-frame outliers appear at high future noise levels,
particularly $\sigma_{\mathrm{future}}=0.9$--$1.0$, with a pronounced peak
around layers 21--23 and a weaker response around layers 16--17.
The peak exceeds an order of magnitude relative to the median spatial-token
norm. At intermediate noise levels these responses weaken, and most lower-noise
settings show much smaller norm disparities. The dependence is not strictly
monotonic: weak localized responses are also visible at zero noise.
These results qualify the observation in Section~\ref{sec:spatial_outliers}:
pretrained Wan2.2 can exhibit first-frame outliers, but their strength depends
on the future-input noise level and network depth. The default
$\sigma_{\mathrm{future}}=1$ used in the main paper lies in the strong-noise
regime where this behavior is most visible. The heatmap does not by itself
establish how frequently the pattern occurs across observations or what
information the outlier tokens encode.

\begin{figure}[H]
    \centering
    \setlength{\tabcolsep}{3pt}
    \begin{tabular}{@{}ccc@{}}
        \small FastWAM & \small FastWAM (retrained A) & \small FastWAM (retrained B) \\
        \multicolumn{3}{@{}l}{\footnotesize Sample 003426} \\[-2pt]
        \includegraphics[width=0.29\linewidth]{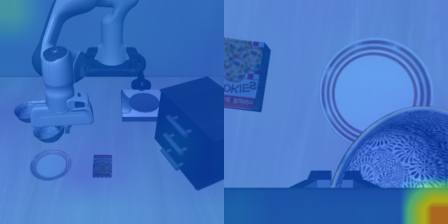} &
        \includegraphics[width=0.29\linewidth]{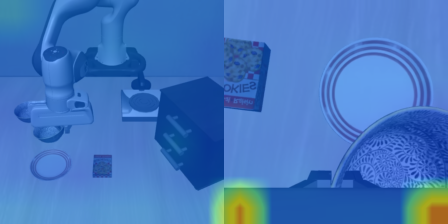} &
        \includegraphics[width=0.29\linewidth]{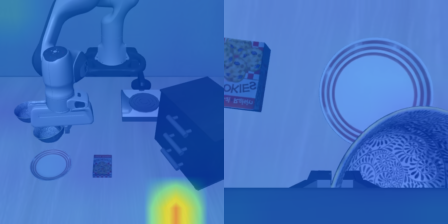} \\[2pt]
        \multicolumn{3}{@{}l}{\footnotesize Sample 050360} \\[-2pt]
        \includegraphics[width=0.29\linewidth]{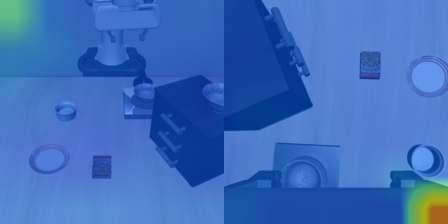} &
        \includegraphics[width=0.29\linewidth]{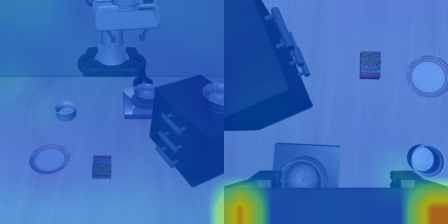} &
        \includegraphics[width=0.29\linewidth]{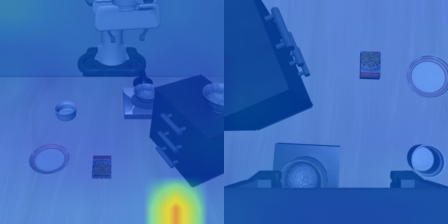} \\[2pt]
        \multicolumn{3}{@{}l}{\footnotesize Sample 226827} \\[-2pt]
        \includegraphics[width=0.29\linewidth]{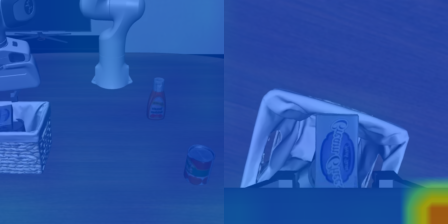} &
        \includegraphics[width=0.29\linewidth]{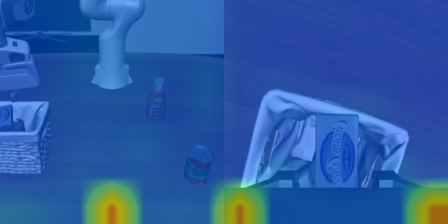} &
        \includegraphics[width=0.29\linewidth]{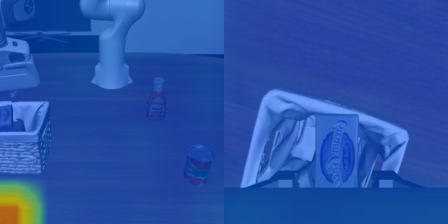}
    \end{tabular}
    \caption{\small\textbf{Additional FastWAM outlier visualizations.}
    Hidden-state norms for three observations in FastWAM and two retrained variants.
    All panels show norms overlaid on the corresponding observations.
    Colors should be compared within each panel.}
    \label{fig:appendix_outlier_visualizations}
    \end{figure}

\paragraph{Additional observations.}
Figure~\ref{fig:appendix_outlier_visualizations} compares hidden-state norms
for three additional observations in FastWAM and two retrained variants.
The maps suggest that each model favors a small set of recurring spatial
locations for high-norm responses, while the active locations and number of
peaks can vary across samples. Thus, these outliers exhibit model-specific
spatial preferences rather than an identical pattern for every observation.

\begin{figure}[H]
    \centering
    {\scriptsize
    \textcolor[HTML]{2E73B7}{\rule{1.2em}{0.7pt}} FastWAM\quad
    \textcolor[HTML]{DF7A18}{\rule{0.5em}{0.7pt}\hspace{0.2em}\rule{0.5em}{0.7pt}} FastWAM + Obs. FM\quad
    \textcolor[HTML]{2A907A}{\rule{0.5em}{0.7pt}\hspace{0.2em}\rule{0.15em}{0.7pt}\hspace{0.2em}\rule{0.3em}{0.7pt}} FastWAM + Reg.\quad
    \textcolor[HTML]{AD4B7D}{\rule{1.2em}{0.7pt}} FastWAM + Reg. + Obs. FM}\par
    \begingroup
    \sbox0{\includegraphics{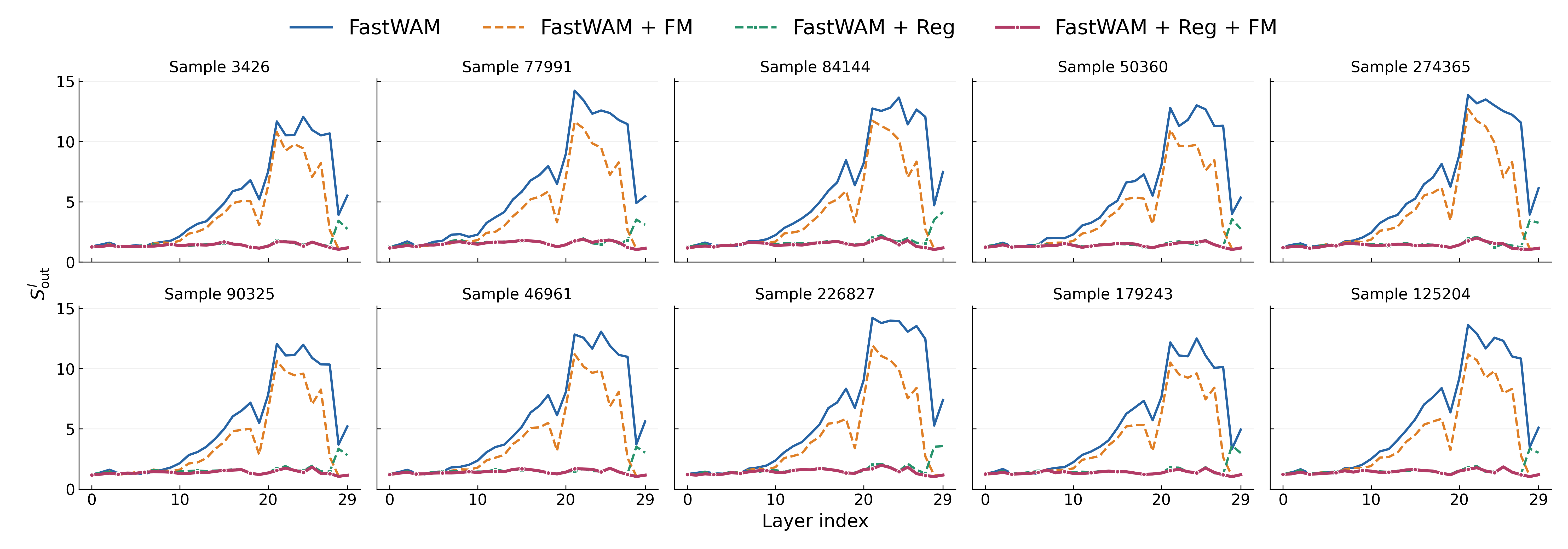}}
    \dimen0=0.09\ht0
    \edef\legendcrop{\the\dimen0}
    \includegraphics[width=\linewidth,trim=0 0 0 \legendcrop,clip]{figures/outliers_analysis/Reg_FM_outlier_curves_10.png}
    \endgroup
    \caption{\small\textbf{Complementary effects of registers and observation FM loss supervision.}
    Layer-wise spatial outlier strength $S_{\mathrm{out}}$ across ten observations.
    Registers suppress intermediate-layer peaks, while observation FM loss supervision
    suppresses outliers in the final layers; together, they suppress
    pronounced peaks throughout the network.}
    \label{fig:reg_fm_outlier_curves}
    \end{figure}

\paragraph{Complementary effects across layers.}
Figure~\ref{fig:reg_fm_outlier_curves} separates the effects of registers and
observation FM loss supervision across ten observations. Registers alone substantially
suppress intermediate-layer spatial outliers, but elevated norms persist in
the final layers. Observation FM loss supervision alone primarily suppresses final-layer
outliers, while pronounced intermediate-layer peaks remain. Combining both
keeps $S_{\mathrm{out}}$ low throughout the network in all ten examples,
indicating complementary effects on spatial outliers at different depths.
\par

\begin{figure}[H]
    \centering
    \begin{minipage}{0.39\linewidth}
        \centering
        {\small Sample 003426}\par\vspace{2pt}
        \includegraphics[width=\linewidth]{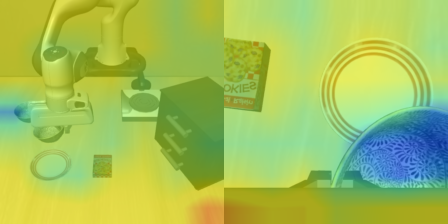}
    \end{minipage}\hspace{0.02\linewidth}%
    \begin{minipage}{0.39\linewidth}
        \centering
        {\small Sample 050360}\par\vspace{2pt}
        \includegraphics[width=\linewidth]{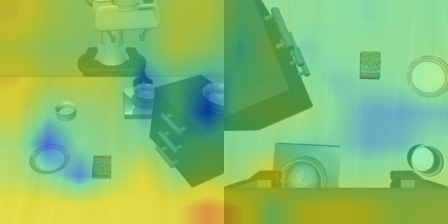}
    \end{minipage}
    \caption{\small\textbf{RegisterWAM spatial norm maps.}
    Hidden-state norms overlaid on two observations, without conspicuous
    localized outliers. Colors are interpreted within each panel.}
    \label{fig:registerwam_clean_norms}
    \end{figure}

\paragraph{Spatial norm maps in RegisterWAM.}
Figure~\ref{fig:registerwam_clean_norms} shows two additional observations
from RegisterWAM. Unlike the isolated high-norm peaks in the FastWAM maps,
these spatial norm maps show broader variations without conspicuous
localized outliers, consistent with the suppression of spatial outliers
when registers and observation FM loss supervision are combined.

\section{Benchmark Details}
\label{app:benchmarks}
Real-world evaluation is detailed below.

\subsection{Real-World Evaluation}
\label{app:real_world}
\paragraph{Evaluation and scoring.}
Each setting is evaluated over 20 trials. We report the mean task score,
which awards partial credit for completed steps, rather than the fraction
of fully successful trials. The overall average in Table~\ref{tab:real}
is the arithmetic mean of the six setting-level scores, including zero-score
settings. RegisterWAM improves all three ID tasks and the height and
distractor variants. Both methods obtain zero score with the blue tablecloth,
indicating a shared failure under this appearance shift in the tested trials.

\paragraph{Task instructions.}
The ID tasks use the following instructions, reproduced from the experiment
record.
\begin{description}
\item[Battery insertion.]
``First, use the left arm. Pick up the two batteries on the left, one at a
time, and insert them into the two left holes. After the left arm finishes,
use the right arm. Pick up the two batteries on the right, one at a time,
and insert them into the two right holes.''
\item[Bowl-and-cup stacking.]
``The robot first grasps a bowl from the table and places it on the table.
It then grasps another bowl and stacks it on top. This stacking action is
repeated once more. After that, the robot grasps a cup from the table and
places it on the table, then grasps another cup and stacks it. Finally,
the robot grasps one more cup and stacks it together with the others.''
\item[Bottle handover.]
``The left arm grasps the white bottle and lifts it into the air, passes it
to the right arm, and the right arm places the white bottle into the orange
plate on the right side of the table.''
\end{description}

\section{Additional Qualitative Results}

\subsection{Visualizing Register Attention}
\label{app:register_visualization}
\begin{figure}[H]
    \centering
    \setlength{\tabcolsep}{4pt}
    \begin{tabular}{@{}cc@{}}
        \includegraphics[width=0.41\linewidth]{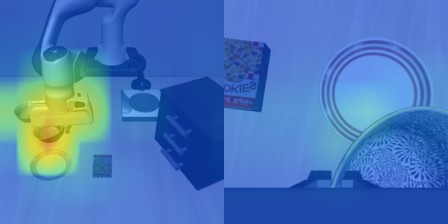} &
        \includegraphics[width=0.41\linewidth]{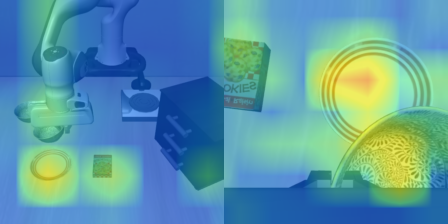} \\[-2pt]
        {\small Reg. 0, Head 3} & {\small Reg. 1, Head 11} \\[5pt]
        \includegraphics[width=0.41\linewidth]{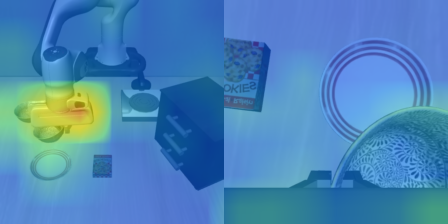} &
        \includegraphics[width=0.41\linewidth]{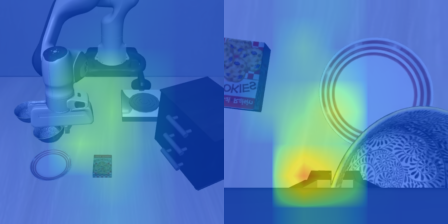} \\[-2pt]
        {\small Reg. 2, Head 16} & {\small Reg. 3, Head 19}
    \end{tabular}
    \caption{\small\textbf{Register attention over spatial tokens.}
    Four selected register--head pairs, overlaid on observation 003426.
    Register and head indices are shown below each panel.}
    \label{fig:register_attention}
    \end{figure}

Figure~\ref{fig:register_attention} visualizes attention from registers to
spatial observation tokens for four selected register--head pairs.
Reg.~0 (Head~3) and Reg.~2 (Head~16) emphasize the gripper region, with
different spatial extents. Reg.~1 (Head~11) distributes attention across
several objects, including the plates on the right, whereas Reg.~3 (Head~19)
emphasizes a lower-right region beside them. These maps illustrate diverse
and partially overlapping spatial attention across register--head pairs;
they do not imply fixed semantic roles for individual registers.

\section{Limitations}
Our evidence supports functional specialization but does not fully establish what information outlier or register tokens encode. The experiments cover a limited family of WAM architectures, and the observed behavior may depend on model scale, training objective, or optimization history.

\end{document}